\def\year{2016}\relax
\documentclass[letterpaper]{article}
\usepackage{aaai16}
\usepackage{times}
\usepackage{helvet}
\usepackage{courier}
\usepackage{marvosym}
\usepackage{graphicx}
\usepackage{amsmath}
\usepackage{amssymb}
\usepackage{algorithmic,algorithm}
\begin{document}
%

\title{DARI: Distance metric And Representation Integration for Person Verification}
\author{Guangrun Wang, Liang Lin\thanks{Corresponding author is Liang Lin. This work is in part supported by China 863 Program (Grant no. 2013AA013801), in part by Guangdong Natural Science Foundation (Grant no. 2014A030313201), in part by Guangdong Science and Technology Program (Grant no. 2013B010406005 and 2015B010128009).}, Shengyong Ding, Ya Li \and Qing Wang\\
School of Data and Computer Science, Sun Yat-sen University, Guangzhou 510006, China\\
 wanggrun@mail2.sysu.edu.cn, linliang@ieee.org, marcding@163.com,\\ liya@gzhu.edu.cn, ericwangqing@gmail.com
}
\maketitle

\begin{abstract}
The past decade has witnessed the rapid development of feature representation learning and distance metric learning, whereas the two steps are often discussed separately. To explore their interaction, this work proposes an end-to-end learning framework called DARI, i.e. Distance metric And Representation Integration, and validates the effectiveness of DARI in the challenging task of person verification. Given the training images annotated with the labels, we first produce a large number of triplet units, and each one contains three images, i.e. one person and the matched/mismatch references. For each triplet unit, the distance disparity between the matched pair and the mismatched pair tends to be maximized. We solve this objective by building a deep architecture of convolutional neural networks. In particular, the Mahalanobis distance matrix is naturally factorized as one top fully-connected layer that is seamlessly integrated with other bottom layers representing the image feature. The image feature and the distance metric can be thus simultaneously optimized via the one-shot backward propagation. On several public datasets, DARI shows very promising performance on re-identifying individuals cross cameras against various challenges, and outperforms other state-of-the-art approaches.
\end{abstract}

\section{Introduction}

\begin{figure}[!htb]
\begin{center}
\includegraphics [width=0.95\linewidth]{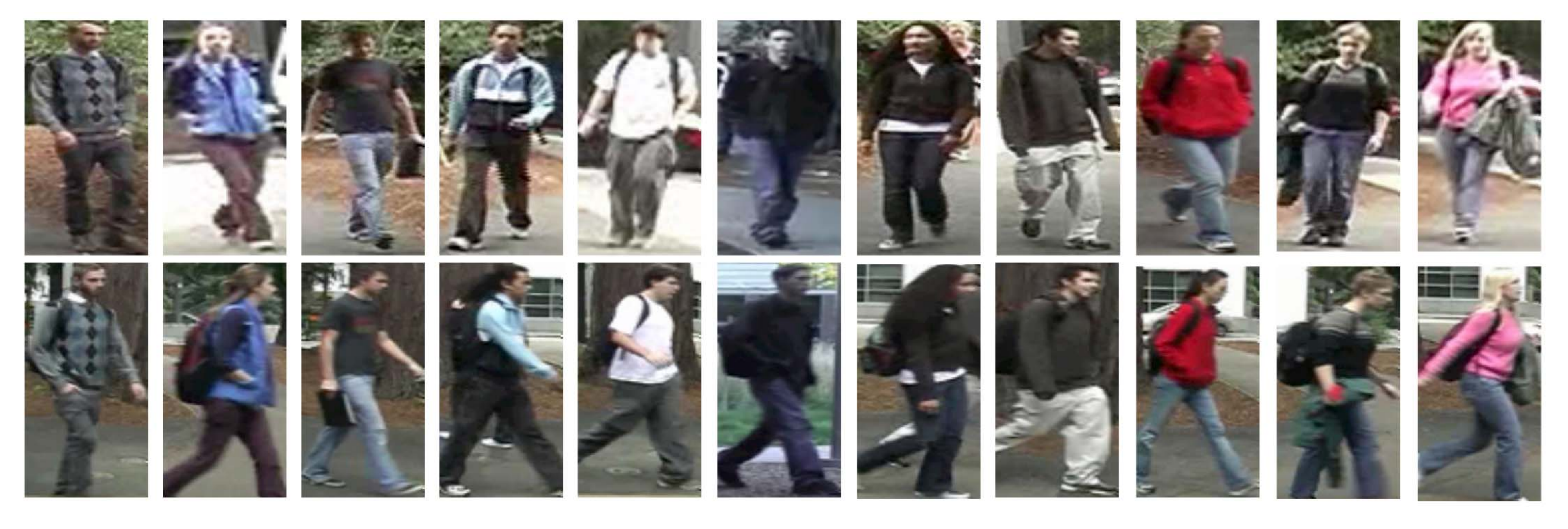}
\caption{Typical examples of person verification across cameras. Each column corresponds to one individual, and the large variations exist between the two examples due to the light, pose and view point changes.}
\label{fig:challenges}
\end{center}
\end{figure}

Distance/similarity measure between images plays a fundamental role in many computer vision applications, e.g., person verification \cite{Hirzer2012re_id},  matching heterogeneous data \cite{Zhai2012multiview}, and multimedia retrieval \cite{Chechik2010image}. Conventional Mahalanobis distance metric learning models, which aim to seek a linear transformation by pulling the distances between similar pairs while pushing the distances between dissimilar pairs, are theoretically appealing in producing discriminative similarity or distance measure from the given training data \cite{weinberger2005distance}. However, these approaches usually are performed in the original data space or the hand-engineered feature space (i.e. representation), and thus are limited in capturing variety of image appearance and handling complicated nonlinear manifold.

In this paper, we investigate the possible interaction between feature learning and distance metric learning, and address the very challenging task of person verification (i.e. matching individuals across cameras). Figure $\ref{fig:challenges}$ shows some examples of this task, where the pedestrians across cameras distinctly vary in appearance and structure caused by pose, lighting and view angle changes. The proposed framework is built based on the convolutional neural network (CNN)~\cite{LeCun1989CNN}, which jointly optimizes the human representation as well as the proper distance metric to robustly match individuals against various real challenges. We call this framework DARI (i.e. Distance metric And Representation Integration).

We aim at preserving similarity of the same person while discriminating the different individuals, and thus define DARI in the form of maximizing relative distance. Specifically, we organize the training images into triplet units, and each unit contains one person image and the matched/mismatch references. For all triplet units, the objective of DARI is to maximize the distance disparity between the matched pairs and the mismatched pairs. In particular, DARI seeks optimal solutions for both feature representation and distance metric, while the existing approaches of person verification~\cite{xu2013human,liu2012person} focuses on only either of the two components. We adopt the deep CNN architecture to extract the discriminative features from the input images, where the convolutional layers, max-pooling operators, and one full connection layer are stacked up. And the Mahalanobis distance matrix is applied with the generated feature as the matching metric. Due to the positive semi-definite requirement for the Mahalanobis metric, directly optimizing the metric matrix is computational intensive. We propose to decompose the Mahalanobis matrix inspired by~\cite{mignon2012pcca}, and further factorize the matrix into a fully-connected layer on the top of our deep architecture. In this way, the distance metric is seamlessly integrated with the image feature represented by the other layers of neural networks. The joint optimization can be then efficiently achieved via the standard backward propagation. Therefore, by means of the nonlinearity learning of deep neural networks, DARI is capable of representing the complicated transformation to identify the people in the wild.

To scale up our approach to the large amount of training data, we implement the training in a batch-process fashion. In each round of training, we randomly select a relatively small number (say $60 \sim 70$) of images, and use them to organize the triplet units. By taking the triplets as the inputs, we update the model parameters by the stochastic gradient descent (SGD) algorithm~\cite{lecun1998gradient}. Another arising issue is that the triplet organization cubically enlarges the number (say $4800$) of training samples, as one image can be included into more than one triplet. To overcome it, we calculate the gradients on the images instead of the produced triplets, and thus reduce the computation cost by making it only depends on the number of the selected images.

The key contribution of this paper is a novel end-to-end framework that naturally fuses the concept of feature learning and metric learning via the deep neural networks. To the best of our knowledge, such an approach is original to the community. On several challenging benchmarks for person verification (e.g., CUHK03 \cite{li2014deepreid},CUHK01\cite{li2012human} and iLIDS \cite{zheng2009people}), our DARI framework demonstrates superior performances over other state-of-the-art approaches.



\section{Related Work}
A number of approaches, e.g., local metric learning and kernelized metric learning, have been suggested to learn multiple or nonlinear metrics from training data with complicated nonlinear manifold structure. In local metric learning, local metrics can be learned independently for each region or by considering the data manifold structure \cite{Noh2010GLML,Wang2012PLML,LMNN}. In kernelized metric learning, a kernel function is exploited to implicitly embed instances into the reproducing kernel Hilbert space (RKHS), and a Mahalanobis distance metric is then learned in the RKHS space \cite{Wang2011MKL}. Actually, kernelized method with Gaussian RBF kernel can also be treated as local learning approach. As pointed out in \cite{Bengio2009Survey}, local learning are also shallow models, and generally are insufficient in coping with highly varying appearance and deformations of images. Another efficient local distance metric learning \cite{yang2006efficient} was also proposed for classification and retrieval. To handle heterogeneous data, \cite{xiong2012random} propose a method  using a random forest-based classifier to strengthen the distance function with implicit pairwise position dependence.

On the other hand, deep convolutional models have been intensively studied and achieved extremely well performance. Compared with the multiple layer perceptron, CNN contains much less parameters to be learned, and can be efficiently trained using stochastic gradient descent. With the increasing of large scale training data and computational resources, deeper CNN and novel regularization methods had been developed, and deep CNN has gained great success in many visual recognition tasks, e.g., image classification \cite{Krizhevsky2012CNN}, object detection \cite{Szegedy2013Detection}, and scene labeling \cite{Pinheiro2014scene}.

Despite the success of deep learning in variety of vision tasks, little studies were conducted on metric learning with deep architecture. Chopra et al. \cite{Chopra2005ML_EBM} suggested a energy-based model (EBM) for discriminative similarity metric learning for image pairs. Stacked restricted Boltzmann machines (RBMs) had also been exploited to learn nonlinear transformation for data visualization and supervised embedding \cite{Min2010dt_RBM}. Cai et al. \cite{Cai2012ISA} proposed a deep nonlinear metric learning method by combining logistic regression and independent subspace analysis. Hu et al. \cite{Hu2014DDML} adopted the forward multi-layer neural network to learn deep metric for hand-crafted features. Compared with these approaches, the proposed DARI model considers the prominence of CNN in capturing salient and incorporates the Mahalanobis distance with the generated image features into one optimization target for distance metric and representation integration.

One approach close to ours was proposed by Wang et al. \cite{wang2014learning}, which addresses the triplet-based similarity learning for image retrieval. However, our work have significant differences with that work. First, we derive our formulation from a novel angle, i.e. integrating feature learning and distance metric learning. Second, our learning method has advantage in the triplet generation and the batch-based gradient descent learning. Specifically, given $m$ training triplets containing $n$ distinct images ($n << m$), their algorithm optimizes with $3\times m$ forward and backward propagations, while only $n$ rounds is required for our approach because we derive to calculate the gradient over the images. Last, our deep architecture is specifically designed (only two conv layers are used) and we train our model from scratch, while they utilized the Alex's model \cite{krizhevsky2012imagenet} that is pre-trained on the ImageNet.

\section{Framework}
\subsection{Primal Formulation}

Given a fixed feature representation, metric learning is to learn a distance function by satisfying the constraint according to the label information of samples. Here we define our formulation via relative distance comparison based on the triplet-based constraint. As is discussed in \cite{ding2015deep}, the triplet models allows the images of one identity lying on a manifold while maximizing the margin between within-class pairs from between-class pairs, and tends to result in better tradeoff between adaptability and discriminability.

More precisely, the relative distance constraint is defined with a set of  triplet units $\mathcal{T} =\{<I_i,I_j,I_k>\}$, in which $<I_i,I_j>$ is a pair of matched images (images of the same individual) and $<I_i,I_k>$ contains two mismatched images from the labeled image set $I=\{I_l, y_l\}$ with $y_l$ denoting the label. Let $\mathbf{M}$ denote the metric matrix and $F_\mathbf{W}(I_i)$ denote the feature representations of the $i$th image learned by the feature network with the network parameters $\mathbf{W}$. Then Mahalanobis distance between $I_i$ and $I_j$ using the CNN features can be written as follows:
\begin{equation}
d^2(I_i,I_j)=\Delta F_\mathbf{W}(I_i,I_j)^T\mathbf{M}\Delta F_\mathbf{W}(I_i,I_j) \label{eq:distance}
\end{equation}
where $\Delta F_\mathbf{W}(I_i,I_j)=F_\mathbf{W}(I_i)-F_\mathbf{W}(I_j)$ denotes the feature difference between the image $I_i$ and $I_j$. For each training triplet $<I_i,I_j,I_k>$ in $\mathcal{T}$, the desired distance should satisfy: $d^2(I_i,I_j)<d^2(I_i,I_k)$. Let $\Delta d^2(I_i,I_j,I_k)$ denote $d^2(I_i,I_k)-d^2(I_i,I_j)$, we turn this relative constraints into the minimization of the following hinge-loss like objective function where $\lambda\operatorname{tr}(\mathbf{M})$ acts as a regularization term as in \cite{shen2012positive}.
\begin{small}
\begin{equation}
\begin{split}
\mathcal{H}(\mathbf{W},\mathbf{M})=&\sum\limits_{\forall  < {I_i},{I_j},{I_k} >  \in \mathcal{T}} {{{(1 - \Delta {d^2}({I_i},{I_j},{I_k}))}_ + }}  + \lambda {\rm{tr}}({\bf{M}})
\end{split}
\nonumber
\end{equation}
\end{small}
In the following, we use $\sum$ to denote $\sum\limits_{\forall  < {I_i},{I_j},{I_k} >  \in \mathcal{T}}$ for notation simplicity. By the definition of $\Delta d^2$, we get the following objective functions:
\begin{small}
\begin{equation}
\begin{split}
\label{equ:overall-loss}
&\mathcal{H}(\mathbf{W},\mathbf{M}) = \sum{}(1 - (\Delta {F_{\bf{W}}}({I_i},{I_k})^T{\bf{M}}\Delta {F_{\bf{W}}}({I_i},{I_k}) \\
&- \Delta {F_{\bf{W}}}({I_i},{I_j})^T{\bf{M}}\Delta {F_{\bf{W}}}({I_i},{I_j})) {)_ + } + \lambda\operatorname{tr}(\mathbf{M})\\
&\mbox{s.t. }\mathbf{M} \succcurlyeq 0
\end{split}
\end{equation}
\end{small}

\begin{figure}[!htb]
\begin{center}
\includegraphics [width=2.8 in]{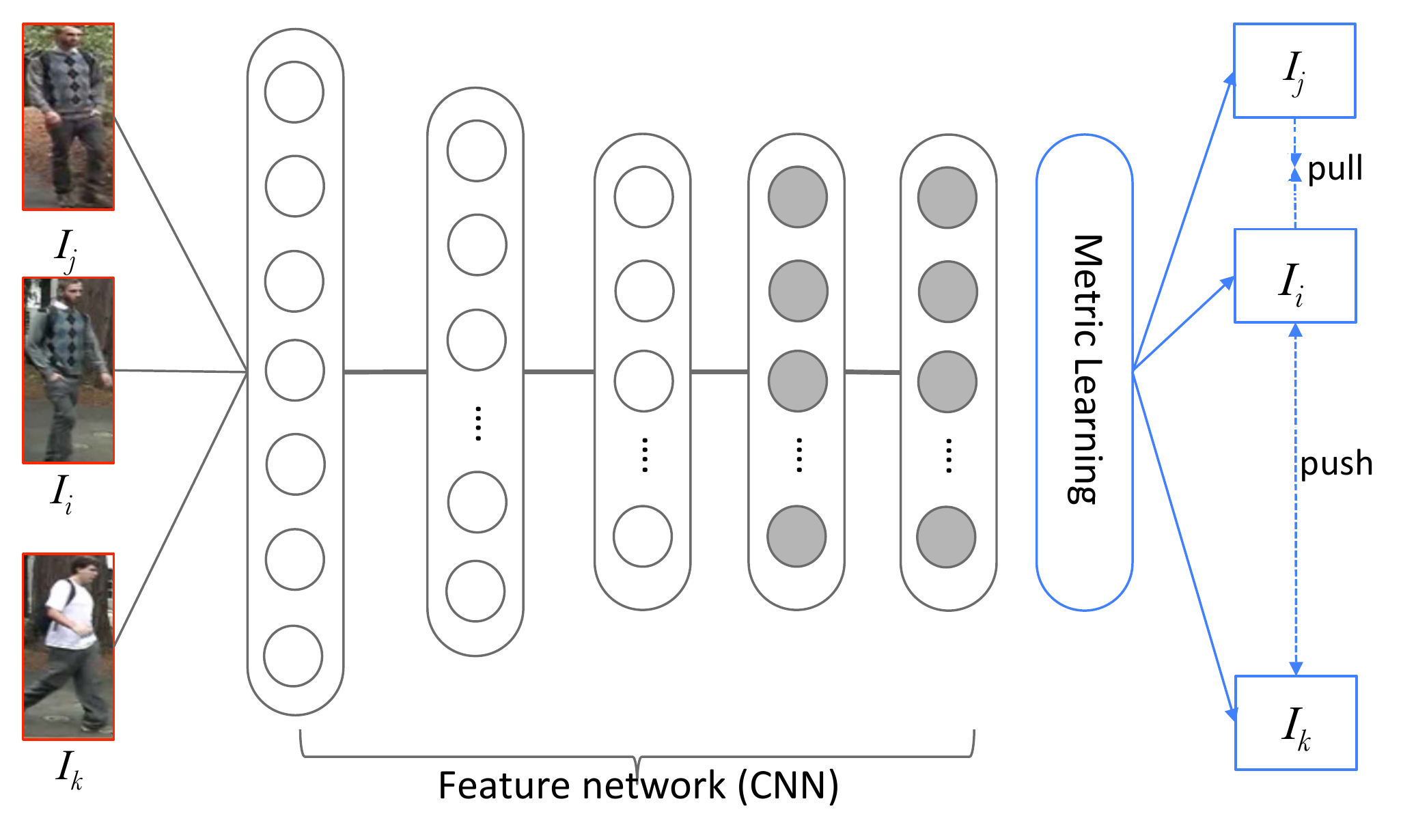}
\caption{Illustration of our learning formulation. It tends to maximize the distance disparity between the matched pair and mismatched pair for each triplet, in terms of optimization. }
\vspace {-3mm}
\label{fig:DARI}
\end{center}
\end{figure}

An intuitive solution to Eqn. (\ref{equ:overall-loss}) is to iteratively optimize $\mathbf{W}$ and $\mathbf{M}$ with either of them fixed. It is, however, computationally expensive, as the PSD projection is necessarily imposed once $\mathbf{M}$ is updated to ensure $\mathbf{M}$ to be positive semi-definite. In this work, to overcome this issue we propose an unified solution by incorporating $\mathbf{M}$ into the deep neural networks.

\subsection{Deep Architecture}

We factorize the metric matrix $\mathbf{M}$ into $\mathbf{L}^T\mathbf{L}$ as $\mathbf{M}$ satisfies the positive semi-definite constraint. The distance measure in Eqn. (\ref{eq:distance}) can be then transformed as,
\begin{small}
\begin{equation}
d^2(I_i,I_j)=||\mathbf{L}\Delta F_\mathbf{W}(I_i,I_j)||^2
\end{equation}
\end{small}

The objective function $\mathcal{H}(\mathbf{W},\mathbf{M})$ in Eqn. (\ref{equ:overall-loss}) can be then reformulated with $\operatorname{tr}(\mathbf{M})=||\mathbf{L}||_{\mathcal{H}}^2$, as
\begin{small}
\begin{equation}
\begin{aligned}
\mathcal{H}(\mathbf{W},\mathbf{L})=&\sum(1 - (||{\bf{L}}\Delta {F_{\bf{W}}}({I_i},{I_k})|{|^2} \\
&- ||{\bf{L}}\Delta {F_{\bf{W}}}({I_i},{I_j})|{|^2}))_ + +\lambda||\mathbf{L}||_{\mathcal{H}}^2 \\
\end{aligned}
\end{equation}
\end{small}

Thus, we can take $\mathbf{M}$ as a linear transform on the output of the CNN-based feature representation. In literature, \cite{weinberger2005distance} \cite{mignon2012pcca} also proposed to decompose the Mahalanobis distance matrix for simplifying the distance metric learning. They attempt to pursue a low-dimensional projection in Euclidean space which embeds the distance metric while ensuring the PSD constraint. However, their solution are complicated requiring additional hypothesis. In this work, we implement a fully connected layer to represent $\mathbf{L}$, which is stacked over the layers representing image features, making the distance metric tightly combined with the deep neural network. Specifically, we treat $\mathbf{L}$ as the neuron weight of the layer, and the network can represent ${\bf{L}}{F_{\bf{W}}}({I_i})$ by taking ${F_{\bf{W}}}({I_i})$ as the input. Then, the feature learning and the distance metric can be thus jointly optimized in an end-to-end way.

In the following, we denote $\mathbf{W}^+=(\mathbf{W},\mathbf{L})$ for notation simplicity. As the regularization term on $\mathbf{L}$ will be automatically implemented by the built-in weight decay  mechanisms in CNN networks, we neglect this part in the objective function.

\begin{small}
\begin{equation}\label{equ:aaa}
\begin{aligned}
\mathcal{H}(\mathbf{W}^+)=&\sum(1 - (||\Delta {F_{{{\bf{W}}^ + }}}({I_i},{I_k})|{|^2}\\
&- ||\Delta {F_{{{\bf{W}}^ + }}}({I_i},{I_j})|{|^2}))_ +  \\
\end{aligned}
\end{equation}
\end{small}

Integrating the metric learning and feature learning into one CNN network yields several advantages. First, this leads to a good property of efficient matching. In particular, for each sample stored in a database, we can precomputed its feature representation and the corresponding decomposed Mahalanobis distance matrix. Then the similarity matching in the testing stage can be very fast. Second, it integrates feature learning and metric learning by building an end-to-end deep architecture of neural networks.

As discussed above, our model defined in Eqn. (\ref{equ:aaa}) jointly handles similarity function learning and feature learning. This integration is achieved by building a deep architecture of convolutional neural networks, which is illustrated in Figure \ref{fig:DARI}. Our deep architecture is composed of two sub-networks: feature learning sub-network and metric learning sub-network. The feature learning sub-network contains two convolution-RELU-pooling layers and one fully-connected layer. Both the pooling layers are max-pooling operations with the size of $3\times 3$ and the stride size is set as $3$ pixels. The first convolutional layer includes $32$ kernels of size $5\times 5\times 3$ with a stride of $2$ pixels. The second convolutional layer contains $32$ filters of size $5\times 5 \times 32$ and the filter stride step is set as $1$ pixel. A fully-connected layer is followed and it outputs a vector of $400$ dimensions. We further normalize the output vector of this fully-connected layer before it is fed to the metric learning sub-network by $y_i=\frac{x_i}{\sqrt{\Sigma x_i^2}}$, where $x_i$, $y_i$ denote the value of the $i$th neuron before and after normalization respectively. Accordingly, the back propagation process accounts for the normalization operation using the chain rule during calculation of the partial derivatives. The metric learning sub-network includes only one fully-connected layer. The neural layer outputs ${\bf{L}}{F_{\bf{W}}}({I_i})$. In this way, the distance metric is tightly integrated with the feature representations, and they can be jointly optimized during the model training.

\section {Learning Algorithm}

Given a labeled dataset with ${M}$ classes (persons) and each class has $N$ images, then the number of all possible meaningful triplets is $N*(N-1)*(M-1)*N*M$. Even for a dataset of moderate size, it is intractable to load all these triplets into the limited memory for the model training. To overcome this issue, we apply batch learning to optimize the parameters, in which the network parameters are updated by the gradient derived only from a small part of all the triplets in each iteration.

\subsection{Batch Process}

In the batch learning process, we need to generate a subset of triplets from all the possible triplets in each iteration. The simplest method is to generate triplets randomly. However, this method makes the number of distinct images be approximately three times the number of the generated triplets because each triplet contains three images, and the likelihood of two triplets sharing the same image is very low. This triplet generation method is very inefficient because there are only a few distance constraints placed on the selected images in each iteration. Instead, to capitalize on the strength of relative distance comparison model, a more reasonable triplet generation method would be one that satisfies the two following conditions:

1. In each iteration, large number of triplets are generated from small number of images to ensure the selected images can be loaded to the memory while rich distance constraints are posed on these images;

2. When increased numbers of iterations are executed, all the possible relative distance constraints between any two classes should be considered in the training process.

These two principles lead to our  proposed triplet generation scheme as follows. In each iteration, we select a fixed number of classes (persons), and construct the triplets only using these selected classes.  More precisely, for each image in each class, we randomly construct a certain number of triplets with the matched reference coming from the same class and the mismatched references coming from the remaining selected classes. The complete mini-batch learning process is presented in Algorithm $\ref{alg:batchTraining}$.

\renewcommand{\algorithmicrequire}{ \textbf{Input:}} 
\renewcommand{\algorithmicensure}{ \textbf{Output:}} 
\begin{small}
\vspace{-3mm}
\begin{algorithm}[htb]
\caption{Learning DARI with batch-process}
\label{alg:batchTraining}
\begin{algorithmic}[1]
\REQUIRE ~~\\
   Training images $\{I_i\}$;
\ENSURE ~~\\
	Network Parameters $\mathbf{W}$
\WHILE    {$t<T$}
\STATE $t\leftarrow t+1$;
\STATE Randomly select a set of classes (persons) from the training set;
\STATE Construct a set of triplets from the selected classes;
\STATE Calculate the gradient $\Delta \mathbf{W}$ for the generated triplets using Algorithm $\ref{alg:gradDescentByImage}$;
\STATE $\mathbf{W}^t=\mathbf{W}^{t-1}-\lambda_t \Delta \mathbf{W}$
\ENDWHILE

\end{algorithmic}
\end{algorithm}
\end{small}

\subsection{Parameter Optimization}

Under the mini-batch training framework, a key step is to calculate the gradient for the triplet set in each iteration. A straight method is to calculate the gradient for each triplet according to the loss function, and sum these gradients to get the overall gradient. But with this approach three separate memory units and a network propagation would be needed for every triplet. This is inefficient as there will be duplicated network propagation for the same image, recalling that for each batch we generate triplets from a known subset of images. We now show that there exists an optimized algorithm in which the computational load mainly depends on the number of distinct images rather than the number of the triplets.

It would be difficult to write the objective function in Eqn. (\ref{equ:aaa}) directly as the sum of image-based loss items because it takes the following form (for notation simplicity, we use $\mathbf{W}$ to denote $\mathbf{W^+}$ in the rest of the paper):
\begin{small}
\begin{equation}
\mathcal{H}(\mathbf{W})=\sum{}loss(F_{\mathbf{W}}(I_i),F_{\mathbf{W}}(I_j),F_{\mathbf{W}}(I_k))\nonumber
\end{equation}
\end{small}
Fortunately, because the loss function for a specific triplet is defined by the outputs of the images in this triplet, the total loss can also be considered as follows, where $\{I'_i\}$ represents the set of all the distinct images in the triplets and $m$ denote the size of the distinct images in the triplets.
\begin{small}
\begin{equation}
\mathcal{H}(\mathbf{W})=\mathcal{H}(F_{\mathbf{W}}(I'_1),F_{\mathbf{W}}(I'_2),...,F_{\mathbf{W}}(I'_i),...,F_{\mathbf{W}}(I'_m)) \nonumber
\end{equation}
\end{small}

By the derivative rule, we have the following equations, where $W^l$ represents the network parameters, $X_i^l$ represents the feature maps of the image $I'_i$ at the $l^{th}$ layer and $\frac{\partial \mathcal{H}}{\partial W^l}(I'_i)$ denote the partial derivative derived from image $I'_i$.
\begin{small}
\begin{equation}
\frac{\partial \mathcal{H}}{\partial W^l}=\Sigma_{i=1}^{m}\frac{\partial \mathcal{H}}{\partial X_i^l}\frac{\partial X_i^l}{\partial W^l}=\Sigma_{i=1}^{m}\frac{\partial \mathcal{H}}{\partial W^l}(I'_i)
\label{equ:paramPartialTriplet}
\end{equation}
\end{small}
\begin{small}
\begin{equation}
\frac{\partial \mathcal{H}}{\partial X_i^l}=\frac{\partial \mathcal{H}}{\partial X_i^{l+1}}\frac{\partial X_i^{l+1}}{\partial X_i^l}
\label{equ:feaPartialTriplet}
\end{equation}
\end{small}
Eqn. \ref{equ:paramPartialTriplet} shows that the overall gradient is the sum of the image-based terms (image-based gradient). Eqn. \ref{equ:feaPartialTriplet} shows that the partial derivative with respect to the feature maps of each image can be calculated recursively. With Eqn. \ref{equ:paramPartialTriplet} and Eqn. \ref{equ:feaPartialTriplet}, the gradients with respect to the network parameters can be obtained by summing the image based gradients using the network back propagation algorithm. The central premise is that we have computed the partial derivative of the output layer's activation for every image, which can be easily obtained from Eqn. \ref{equ:aaa}. Algorithm $\ref{alg:gradDescentByImage}$ gives the detailed process. This optimized algorithm has two obvious merits:

1. We can conveniently use exiting deep learning implementations such as Caffe\footnote{http://caffe.berkeleyvision.org/} to train our model.

2. The number of network propagation executions can be reduced to the number of distinct images in the triplets, a crucial advantage for large scale datasets.
\renewcommand{\algorithmicrequire}{ \textbf{Input:}} 
\renewcommand{\algorithmicensure}{ \textbf{Output:}} 
\begin{small}
\vspace{-3mm}
\begin{algorithm}[htb]
\caption{Calculating gradients for optimization}
\label{alg:gradDescentByImage}
\begin{algorithmic}[1]
\REQUIRE ~~\\
   Training triplets $\mathcal{T}=\{<I_i,I_j,I_k>\}$;
\ENSURE ~~\\
	The gradient of network parameters: $\Delta \mathbf{W}=\frac{\partial \mathcal{H}}{\partial \mathbf{W}}$
\STATE Collect all the distinct images $\{I'_i\}$ in $\mathcal{T}$
\FORALL {$I'_i$}
   \STATE Calculate $F_\mathbf{W}(I'_i)$ by forward propagation;
\ENDFOR
\FORALL {$I'_i$}
\STATE $partialSum=0$;
\FORALL {triplet $<I_i,I_j,I_k>$}
	 \IF {$\Delta d^2(I_i,I_j,I_k)<1$}
        \IF {$I'_i$=$I_i$}
            \STATE $partialSum+=2(F_\mathbf{W}(I_k)-F_\mathbf{W}(I_j))$
        \ELSIF {$I'_i$=$I_j$}
           \STATE $partialSum-=2(F_\mathbf{W}(I_i)-F_\mathbf{W}(I_j)$
        \ELSIF {$I'_i$=$I_k$}
           \STATE $partialSum+=2(F_\mathbf{W}(I_i)-F_\mathbf{W}(I_k)$
        \ENDIF
        \ENDIF
\ENDFOR
\STATE Set the partial derivative with respect to the outputs using $partialSum$
\STATE Calculate $\frac{\partial \mathcal{H}}{\partial \mathbf{W}}(I'_i)$ using back propagation;
\STATE Sum the partial derivative: $\Delta \mathbf{W}$+=$\frac{\partial \mathcal{H}}{\partial \mathbf{W}}(I'_i)$;

\ENDFOR
\end{algorithmic}
\end{algorithm}
\vspace{-3mm}
\end{small}
\section{Evaluations}

{\em Datasets and Implementation details.} We conduct our experiments using three challenging human verification datasets, i.e. CUHK03\cite{li2014deepreid}, CUHK01\cite{li2012human} and iLIDS\cite{zheng2009people} . All the images are resized to 250 $\times$ 100 for the experiment. The weights of the filters and the full connection parameters are initialized from two zero-mean Gaussian distributions with standard deviation 0.01 and 0.001 respectively. The bias terms were set with the constant 0. During the training, we select 60 persons to construct 4800 triplets in each iteration. Before feeding to the network, the images are mirrored with 0.5 probability and cropped to the size 230 $\times$ 80 at the center with a small random perturbation to augment the training data.  We implement our learning algorithm based on the Caffe framework, where we revise the data layer and loss layer to generate the triplets and apply our loss function. We execute the code on a PC with GTX780 GPU and quad-core CPU. And stop the training process when there are less than 10 triplets whose distance constraints are violated, i.e. the distance between the matched pair is greater than the distance between the mismatched pair.

{\em Evaluation Protocol.} We adopt the widely used cumulative match curve (CMC) approach \cite{gray2007evaluating} for quantitative evaluation. We follow the standard setting to randomly partition each dataset into training set and test set without overlap. In each testing, the test set is further divided into a gallery set and a probe set without overlap for 10 times. A rank $n$ rate is obtained for each time, and we use the average rate as the final result.

{\em Component analysis.} In order to demonstrate how the joint optimization of distance metric with feature representation contributes to performance, we implement a simplified model for comparison by discarding the distance metric learning(i.e. the last neural layer). In this implementation, we only optimize CNN-based feature representation by the back-propagation method.


\textbf{Experiments on CUHK03 Dataset.} This benchmark \cite{li2014deepreid} is  the largest one up to date, which contains 14096 images of 1467 pedestrians collected from 5 different pairs of camera views, making it an ideal place for deep learning. Each person is observed by two disjoint camera views and has an average of 4.8 images in each view. We follow the standard setting of using CUHK03 to randomly partition this dataset for 10 times without overlap, and a training set (including 1367 persons) and a test set (including 100 persons) are obtained. In each testing, the testing set is further randomly divided into a gallery set of 100 images (i.e. one image per person) and a probe set (including images of individuals from different camera views in contrast to the gallery set) without overlap for 10 times.

\begin{figure*}
  \centering
  \includegraphics[width=0.75\textwidth]{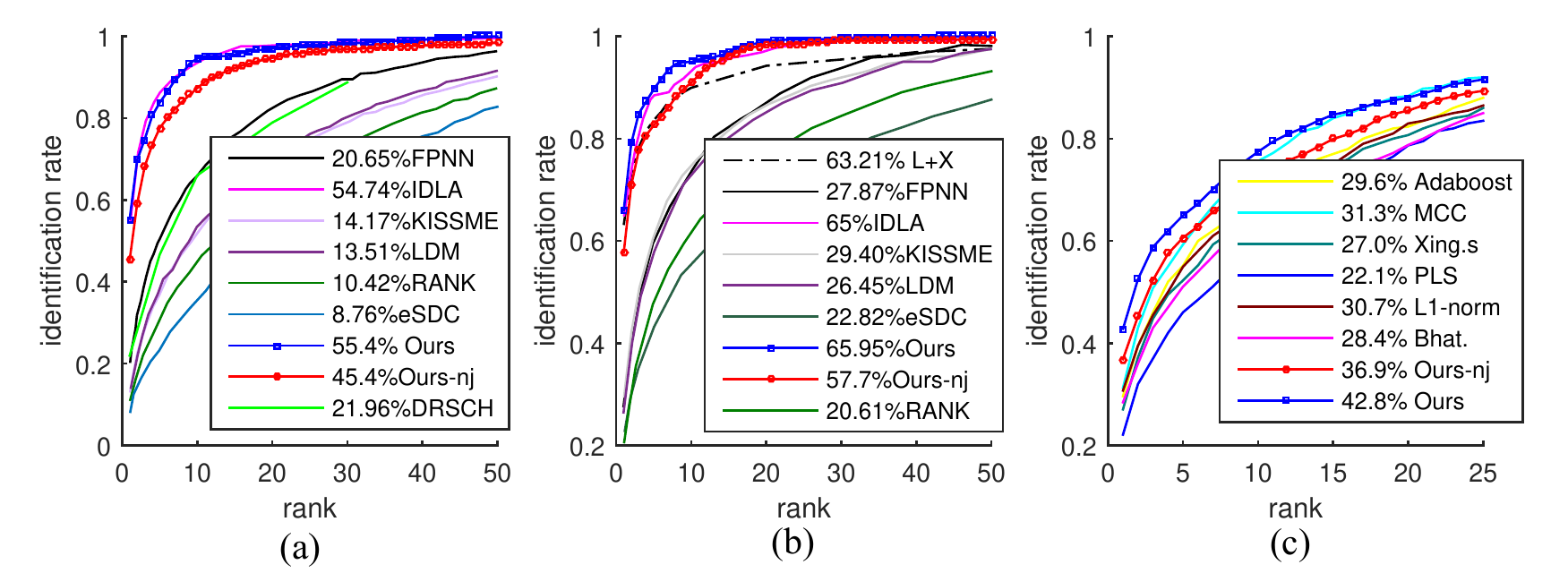}
 \vspace{-3mm}
  \caption{Quantitative results on the three datasets: (a) CUHK03, (b) CUHK01 and (c) iLIDS dataset. Our DARI framework leads superior performances over existing state-of-the-arts overall. Note that ``Ours-nj'' represents a simplified version of our model, i.e. discarding the joint optimization of distance metric and CNN-based feature. }
\label{fig:comMetricLearning}       
\vspace{-3mm}
\end{figure*}

We introduce several types of state-of-the-arts in this experiment. First, we adopt Local Distance Metric Learning (LDM) \cite{guillaumin2009you}, the learning-to-rank method (RANK) \cite{mcfee2010metric} for comparison, which learn distance metrics based on a fixed feature representation. Following their implementation, the handcrafted features of dense color histograms and dense SIFT uniformly sampled from patches are adopted. Two methods especially designed for person re-identification are introduced in this experiment: KISSME \cite{kostinger2012large}, eSDC \cite{zhao2013unsupervised}. Moreover, we compare with a recently proposed deep learning method, DRSCH \cite{zhang2015bit}, FPNN \cite{li2014deepreid} and IDLA \cite{ahmed2015improved}. DRSCH \cite{zhang2015bit} learns hashing code with regularized similarity for image retrieval and person re-identification. FPNN \cite{li2014deepreid} learns pairs of filters to extract person representation and IDLA \cite{ahmed2015improved} is also recently proposed deep learning method for person re-identification.

The results are shown in Fig. \ref{fig:comMetricLearning} (a). It is encouraging to observe that our approach achieves a new state-of-the-art on CUHK03. Note that without the joint optimization of distance metric and representation, the performance (i.e., `` Ours-nj'') degenerates from 55.4\% to 45.4\%.

\textbf{Experiments on CUHK01 Dataset.}

CUHK01 contains 971 individuals, each of which has two samples captured by two disjoint camera views. We partition this dataset into a training set and a testing set exactly following \cite{li2014deepreid}\cite{ahmed2015improved}:  100 persons are used for testing and the remaining 871 persons for training.  Each person has two images for each view and we randomly select one into the gallery set. Single-shot is adopted in the evaluation of CMC curve.

In addition to comparing with the methods adopted in the experiment on CUHK03, we introduce a recently proposed method which also addresses the interaction of representation learning and metric Learning (denoted as L + X) \cite{liao2015person}.

Fig. \ref{fig:comMetricLearning} (b) shows the comparison of our DARI framework with other approaches. DARI achieves a new state of the art, with a rank-1 recognition rate of 65.95\%. The gain of the joint optimization of distance metric and CNN-based feature is also clear on this dataset, 65.95\% over 57.7\%.

\textbf{Cross-dataset Evaluation}

The iLIDS dataset \cite{zheng2009people} was constructed from video images captured in a busy airport arrival hall. It has 119 pedestrians, totaling 479 images. We conduct a challenging task, i.e. cross-dataset task using this dataset, which accords with the real-world surveillance applications. Specifically, we randomly split this dataset into a gallery set and a probe set: the gallery contains only one image of each pedestrian and the remaining images form the probe set. Our model is trained on CUHK03 and tested on this iLIDS dataset without fine-tuning the parameters.

We compare our DARI with several existing methods such as Xing's \cite{xing2002distance}, and MCC \cite{globerson2005metric}. They all use an ensemble of color histograms and texture histograms as the feature representation. Note that the results reported by these competing methods are generated by a different setting: both of the training and the testing data are from this dataset.

Fig. \ref{fig:comMetricLearning} (c) shows the quantitative results. Our superior performance over other approaches demonstrate the good generalization power of DARI. On the other hand, without incorporating Mahalanobis distance matrix, the performance (i.e. ``Ours-nj'' in ) clearly degenerates from 42.8\% to 36.9\%, which highlights the significance of the joint optimization of feature representation and distance metric. In the following, we further evaluate our approach under different implementation setting on iLIDS.




\textbf{Data Augmentation Strategy}. We crop the center of the images with random perturbation to augment the training data. This mechanism can effectively alleviate the over-fitting problems. Without this augmentation scheme, the top 1 performance drop by about 30 percent relatively.

\textbf{Triplet Generation Scheme}. We compared two generation strategy. In the first strategy, we select 60 persons for each iteration and only construct 60 triplets for these persons. In the second strategy, we select the same number of persons while constructing 4800 triplets for these persons. As expected by our analysis, the learning process of the first strategy is much slower than the second strategy and when the learning process of the second strategy converges in 7000 iterations, the performance of the first strategy only achieves about 70 percent of the second strategy.

\section{Conclusion}

We have presented a novel deep learning framework incorporating Mahalanobis distance matrix with convolutional neural networks.  In future work, we will extend our approach for larger scale heterogeneous data, thereby exploring new applications.

\bibliographystyle{aaai}
\bibliography{egbib}

\end{document}